# PhyDCM: A Reproducible Open-Source Framework for AI-Assisted Brain Tumor Classification from Multi-Sequence MRI

**Hayder Saad Abdulbaqi[1], Mohammed Hadi Rahim[1], Mohammed Hassan Hadi[1], Haider Ali Aboud[1], Ali Hussein Allawi[1]**

[1]*Department of medical physics*
*College of Science, University of Al-Qadisiyah*
*Address : Diwaniyah, Iraq*
*E-mail:* hayder.abdulbaqi@qu.edu.iq

*Abstract:* MRI-based medical imaging has become indispensable in modern clinical diagnosis, particularly for brain tumor detection. However, the rapid growth in data volume poses challenges for conventional diagnostic approaches. Although deep learning has shown strong performance in automated classification, many existing solutions are confined to closed technical architectures, limiting reproducibility and further academic development. PhyDCM is introduced as an open-source software framework that integrates a hybrid classification architecture based on MedViT with standardized DICOM processing and an interactive desktop visualization interface. The system is designed as a modular digital library that separates computational logic from the graphical interface, allowing independent modification and extension of components. Standardized preprocessing, including intensity rescaling and limited data augmentation, ensures consistency across varying MRI acquisition settings. Experimental evaluation on MRI datasets from BRISC2025 and curated Kaggle collections (FigShare, SARTAJ, and Br35H) demonstrates stable diagnostic performance, achieving over 93% classification accuracy across categories. The framework supports structured, exportable outputs and multi-planar reconstruction of volumetric data. By emphasizing transparency, modularity, and accessibility, PhyDCM provides a practical foundation for reproducible AI-driven medical image analysis, with flexibility for future integration of additional imaging modalities.

## Introduction

Medical imaging has evolved into one of the foundational pillars of modern disease diagnosis, offering clinicians a non-invasive window into anatomical and functional structures within the human body. Among the diverse modalities available today, magnetic resonance imaging (MRI), computed tomography (CT), and positron emission tomography (PET) each exploit distinct physical phenomena to generate tissue contrast, and each contributes unique diagnostic information. The continued technological advancement of these modalities has led to a substantial increase in both the volume and complexity of generated data, introducing new challenges related to analysis mechanisms, interpretive accuracy, and the speed at which appropriate diagnostic conclusions can be reached [1], [2].

Within this expanding landscape, brain tumor diagnosis represents a particularly demanding application. The subtle morphological variations distinguishing gliomas, meningiomas, and pituitary tumors from normal tissue require careful visual assessment, a task that becomes increasingly burdensome as caseloads grow. Reliance on traditional human interpretation alone is no longer sufficient to keep pace with the rising volume of MRI examinations, especially in cases requiring precise analysis of complex details or subtle pathological patterns. Consequently, growing attention has been directed toward utilizing artificial intelligence techniques, particularly deep learning models, to support medical image analysis and improve diagnostic efficiency while reducing the cognitive burden on specialists [3], [4].

Considerable progress has been achieved in applying convolutional neural networks and, more recently, transformer-based architectures to brain tumor classification from MRI data. However, despite this progress, many existing intelligent solutions still lack sufficient flexibility in terms of adaptability and extensibility within the medical domain. Their dependence on proprietary or monolithic codebases restricts applicability in academic and research contexts, where transparency, reproducibility, and the ability to modify system components are fundamental requirements. Moreover, the unclear separation between theoretical and implementation aspects in some systems makes them difficult to study or systematically develop [5], [6]. An equally important consideration is the disconnect between algorithmic capability and practical deployment infrastructure. A well-performing classification model, taken in isolation, addresses only one facet of the diagnostic workflow. In practice, clinical images arrive in standardized DICOM format, carry extensive metadata governing their interpretation, and require preprocessing pipelines that account for modality-specific intensity characteristics. Few existing open-source tools provide an integrated pathway from raw DICOM data through preprocessing to AI-based inference and structured result reporting within a single, cohesive framework [7], [8].

Based on these considerations, this work presents PhyDCM, a structured and practical framework that integrates artificial intelligence techniques with medical image processing and diagnosis within a flexible, open-source software library. The framework focuses on brain tumor classification from MRI data using a MedViT-based hybrid architecture that combines convolutional feature extraction with transformer-based attention mechanisms. Rather than proposing merely a neural network, the contribution encompasses a modular Python library, standardized preprocessing pipelines, and an interactive desktop application supporting multi-planar reconstruction and structured diagnostic output. The principal contributions of this work can be summarized as follows. First, PhyDCM provides an open-source diagnostic library that cleanly separates inference logic from user interface concerns, enabling independent development and evaluation of each component. Second, the framework integrates DICOM handling, MRI preprocessing, and deep learning inference within a unified pipeline that preserves metadata traceability. Third, a PyQt5-based desktop application offers multi-planar visualization and asynchronous diagnostic execution, bridging the gap between algorithmic research and practical usability. Fourth, experimental validation on curated MRI brain tumor datasets demonstrates stable classification performance, and external validation on independent datasets confirms the generalizability of the approach. Although the current experiments focus on MRI, the modular architecture is designed to accommodate future extension to additional imaging modalities such as CT and PET.

## Related Work
### Deep Learning for Brain Tumor MRI Classification
The application of deep learning to brain tumor classification from MRI data has progressed rapidly over the past decade. Early approaches relied on conventional convolutional neural network architectures, with studies by Abiwinanda et al. [9] and Swati et al. [10] demonstrating that transfer learning from ImageNet-pretrained models could yield competitive performance even with relatively modest medical imaging datasets. The residual network architecture introduced by He et al. [11] proved particularly influential, as its skip connections addressed the vanishing gradient problem that had previously limited the training of very deep networks. More recent investigations have explored architectures specifically adapted to the characteristics of medical images. Deepak and Ameer [12] evaluated GoogLeNet for brain tumor classification and reported that domain-specific fine-tuning substantially improved accuracy compared to off-the-shelf feature extraction. Cheng et al. [13] proposed augmented datasets and region-based analysis to address the challenge of limited training samples. The U-Net architecture developed by Ronneberger et al. [14] established itself as a standard reference for biomedical segmentation, demonstrating the value of encoder-decoder structures with skip connections for dense prediction tasks.

The introduction of vision transformers by Dosovitskiy et al. [15] represented a paradigm shift in image recognition, replacing local convolutional operations with global self-attention mechanisms. Subsequent work adapted this concept for medical imaging. The MedViT architecture proposed by Manzari et al. [16] exemplifies hybrid designs that combine convolutional stages for efficient local feature extraction with transformer blocks for global context modeling. Similarly, Hatamizadeh et al. [17] introduced UNETR, merging transformer encoders with U-Net decoders for volumetric medical image segmentation.

Despite these advances, a critical review of the literature reveals that most classification studies report results using a single dataset split, often without external validation on independent data. Litjens et al. [18] emphasized this concern in their comprehensive survey, noting that generalization across different scanner manufacturers and acquisition protocols remains insufficiently addressed. This observation motivates the inclusion of cross-dataset evaluation in the present work.

### *AI-Based Medical Imaging Frameworks*

Beyond individual model architectures, the broader ecosystem of software tools for medical image analysis has received increasing scholarly attention. The MONAI framework [19], built upon PyTorch, provides domain-specific transforms, data loading utilities, and reference implementations tailored to medical imaging research. Its design philosophy emphasizes reproducibility and adherence to best practices established through extensive community experience. TorchIO [20] offers PyTorch-native augmentation and loading utilities that preserve spatial metadata essential for volumetric medical images. At the data handling level, pydicom [21] serves as the foundational library for parsing DICOM files in Python, providing standards-compliant metadata access with minimal external dependencies. Visualization platforms such as 3D Slicer [22] and ITK-SNAP [23] address the annotation and inspection requirements of medical image analysis workflows. However, these tools typically operate as standalone applications rather than components within integrated diagnostic pipelines. Topol [24] observed that the convergence of human and artificial intelligence in medicine necessitates not only algorithmic performance but also transparent, interpretable, and reproducible systems. From a research perspective, the absence of integrated frameworks combining DICOM processing, AI inference, and interactive visualization within a single open-source library represents a notable gap. Commercial solutions exist but restrict academic use, while existing open-source tools address individual components without providing end-to-end integration.

### *Research Gap*

A synthesis of the reviewed literature identifies several persistent challenges. First, many classification studies rely on rigid experimental setups that resist modification by other researchers, limiting independent verification and extension. Second, the practical requirements of medical image analysis, including DICOM parsing, metadata handling, and interactive visualization, are often treated as peripheral concerns rather than integral components of the diagnostic workflow. Third, the separation between model training codebases and deployment interfaces introduces inconsistencies that may compromise reproducibility. The PhyDCM framework responds to these challenges by providing an integrated, modular, and open-source solution that treats the entire diagnostic pipeline as a coherent system.

## Materials and methods
### PhyDCM Framework Overview
The PhyDCM system consists of two principal components: the PhyDCM Python library and the PhyDCM desktop application. The library serves as the backend intelligence layer, encapsulating all computational logic for medical image preprocessing, deep learning model management, and inference execution. The desktop application functions as a lightweight interaction layer that delegates computational tasks to the library while providing medical image visualization, user input management, and structured result presentation. This architectural separation ensures that the library can be reused independently in command-line workflows, scripting environments, or alternative interfaces.

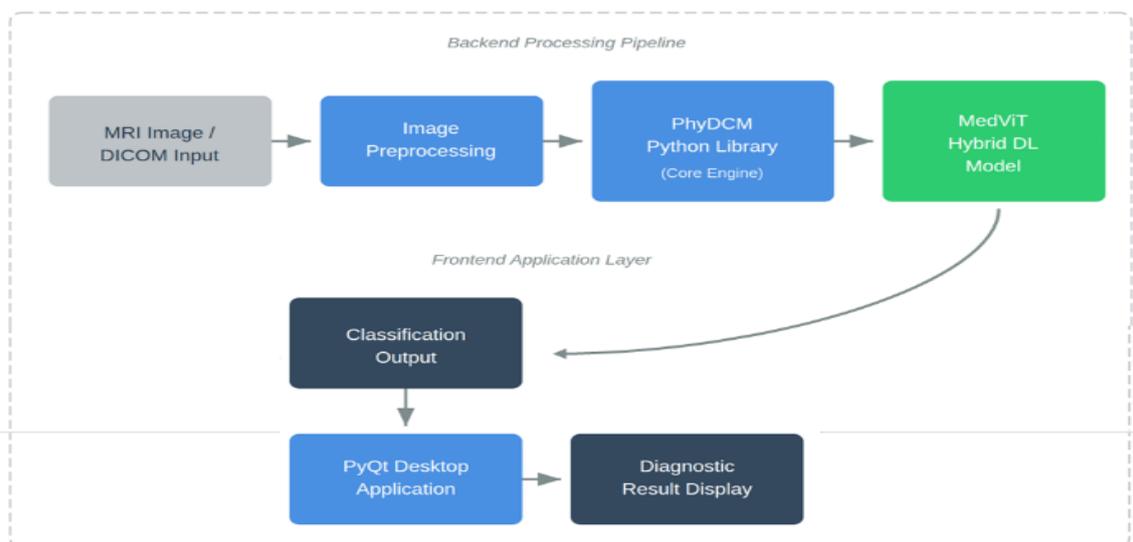

Fig. 1. Overall architecture of the PhyDCM framework showing the separation between the Python library (backend intelligence layer) and the desktop application (user interaction layer).

### Integration of AI in the Diagnostic Workflow
The integration of deep learning inference within the diagnostic workflow follows a clearly defined sequence. Upon receiving an image input and a user-specified scan type, the system activates the corresponding trained model through an automatic binding mechanism. The predictor module scans a designated directory for available models and their associated label mappings, loading each pair dynamically without requiring manual path configuration. This design simplifies model management and supports seamless extension to additional modalities by placing new model files in the appropriate directory structure. The inference pipeline handles both standard image formats and DICOM files, applying modality-appropriate preprocessing before passing normalized arrays to the loaded model. Diagnostic outputs are compiled into structured records that encapsulate predicted categories, confidence scores, image quality metrics, and processing timestamps. This consistent output structure ensures compatibility with downstream operations including tabular display, CSV export, and JSON-based history management.

### *PhyDCM Python Library Design*
The PhyDCM library is designed as a standalone computational module that can operate independently of the desktop application. Its primary objective is to provide a unified and extensible framework for processing medical images and performing deep learning-based inference in a consistent and reproducible manner. By encapsulating preprocessing, model management, and inference within a dedicated library, the system enables researchers to focus on algorithm development and experimentation without being tightly coupled to a specific graphical interface.

### *Library Architecture*
The library is organized into clearly separated modules: configuration management (config.py), dataset creation and loading (create_datasets.py), model construction (medvit_model.py), training orchestration (train.py), inference execution (predict.py), and utility functions (utils.py). Configuration parameters are centralized to ensure consistency across training and inference stages. The model construction module encapsulates the MedViT architecture within a dedicated function, enabling consistent instantiation across different modalities without code duplication.

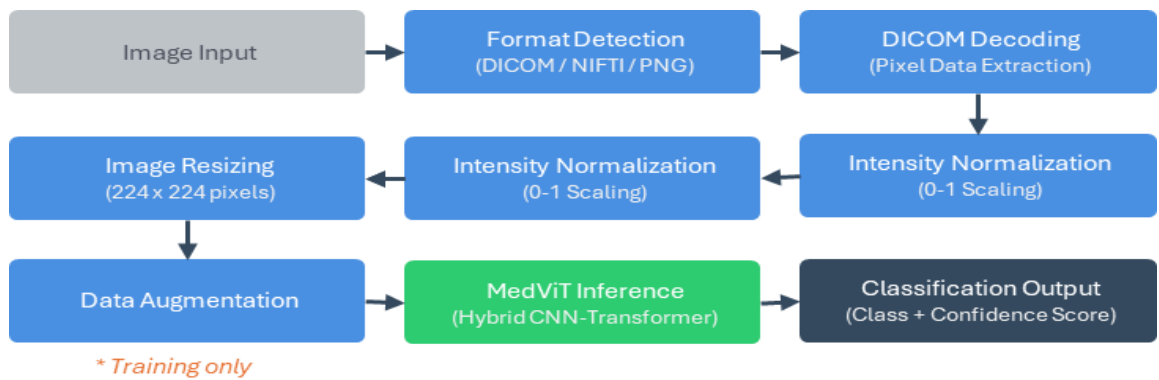

Fig. 2. Internal workflow of the PhyDCM Python library illustrating data flow from DICOM input through preprocessing, model inference, and structured output generation.

### *Core Modules*
The predictor module implements automatic model loading by scanning the designated models directory for trained weight files and corresponding label mappings. When instantiated with a specific scan type filter, only the relevant modality model is loaded, reducing memory overhead. The prediction function accepts file paths, handles format detection, applies appropriate preprocessing, and returns structured results including the predicted class label, confidence score, and probability distribution across all categories.

The training module orchestrates the complete training pipeline: loading datasets through Keras data generators, constructing the MedViT model with modality-specific output dimensions, configuring callbacks for checkpointing and learning rate adjustment, executing the training loop, and saving the best-performing model. Each modality maintains independent training configuration, reflecting the distinct image characteristics associated with different acquisition techniques.

## PhyDCM Desktop Application Architecture

The desktop application is implemented using PyQt5 and structured around a central execution script. The interface is divided into functional regions: a left panel for patient metadata input and diagnostic controls, a central region hosting four image viewers for multi-planar visualization, and a right panel displaying diagnostic results in tabular form. This layout reflects typical medical imaging workflows where image visualization, parameter input, and diagnostic output are accessed simultaneously.

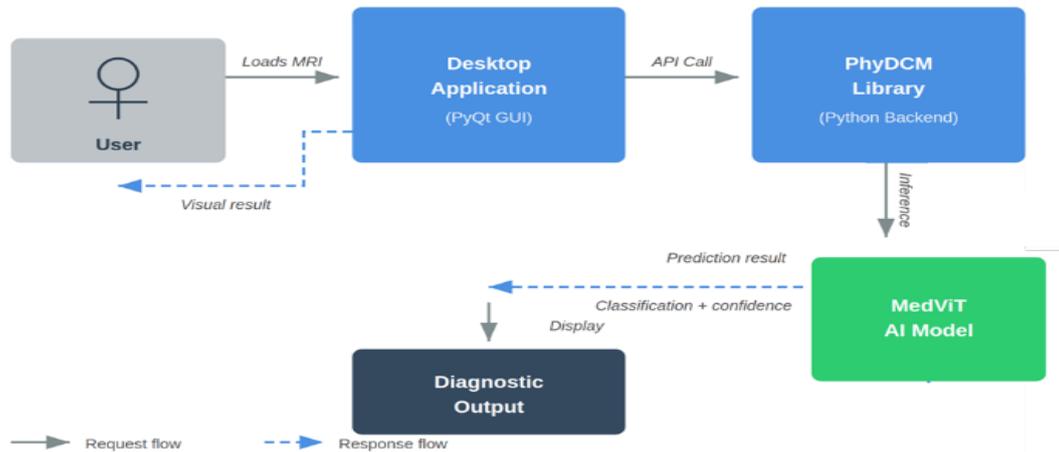

Fig. 3. Functional workflow of the desktop application from image loading through diagnostic execution to structured result export.

## Graphical User Interface

The interface supports theme customization between light and dark modes, accommodating different viewing preferences and reducing visual fatigue during extended usage. An integrated command terminal provides real-time feedback on internal operations, supporting transparency during development and educational demonstrations. The help system guides users through application functionality without requiring external documentation.

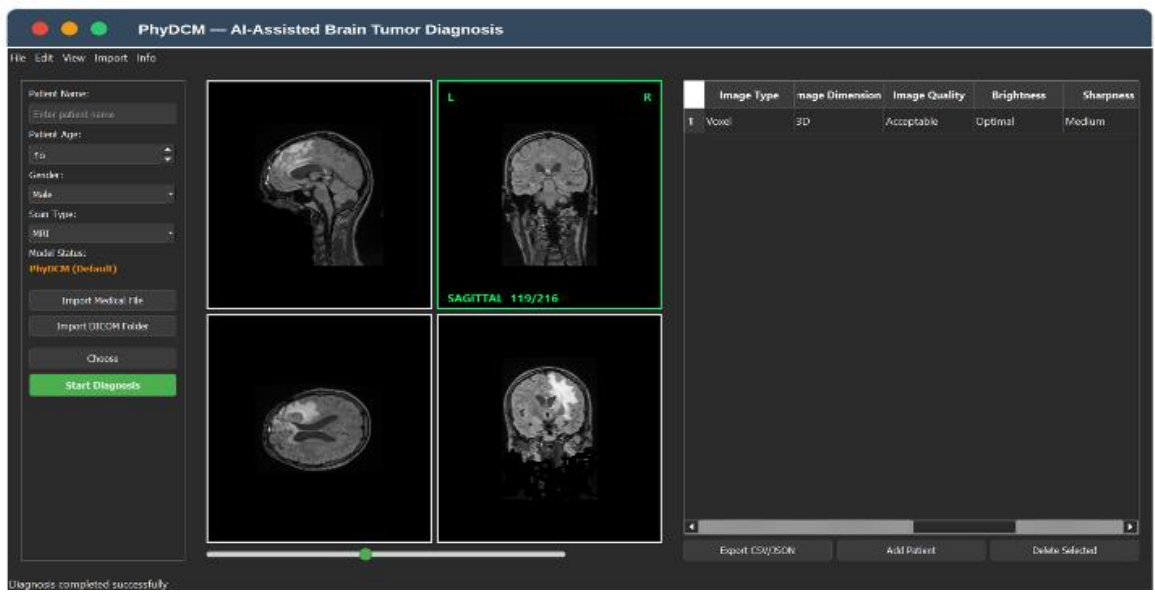

Fig. 4. Graphical user interface of the PhyDCM desktop application showing multi-planar MRI visualization, diagnostic controls, and structured results table.

*MRI Visualization*
Multi-planar reconstruction enables users to explore volumetric DICOM data across axial, sagittal, and coronal planes. DICOM series are loaded and assembled into coherent three-dimensional volumes by ordering individual slices based on acquisition metadata. Interactive slice navigation supports both coarse and fine-grained exploration, with a dedicated selection mechanism enabling synchronized triple-view linking where a point selected in one plane is reflected across the remaining planes. Active view highlighting provides visual feedback regarding the currently focused plane, reducing ambiguity during multi-view interaction.

*MRI Datasets and Preprocessing*
*MRI Datasets*
The experimental evaluation employed curated brain tumor MRI datasets organized into four diagnostic categories: glioma, meningioma, pituitary tumor, and no tumor. The primary training dataset was assembled from multiple complementary sources, including collections provided through academic supervision originating from prior doctoral research and publicly accessible repositories distributed through open research platforms. All images were reorganized according to diagnostic category using directory-based organization to facilitate seamless integration with data loading utilities.

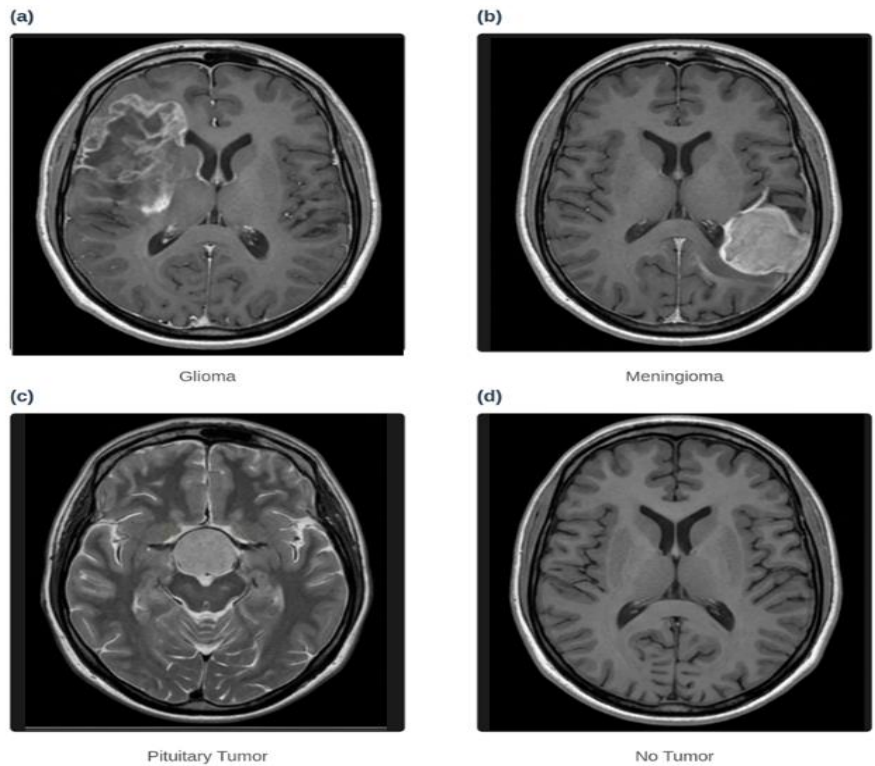

Fig. 5. Representative MRI samples from the four diagnostic categories: (a) glioma, (b) meningioma, (c) pituitary tumor, and (d) no tumor.

To assess generalization capability beyond the primary training distribution, two independent external validation datasets were employed: the Nickparvar Brain Tumor MRI Dataset and the Br35H Brain Tumor Detection dataset. These datasets were used exclusively for evaluation without any contribution to model training, providing an unbiased assessment of cross-dataset transferability.

*Data Preparation*

Due to the heterogeneity of data sources, significant effort was invested in standardizing the dataset structure. Directory-based organization was adopted where each diagnostic category is represented by a dedicated folder within its corresponding modality. This structure enables clear separation between training and validation samples and ensures reproducibility across experimental runs.

*Preprocessing Pipeline*

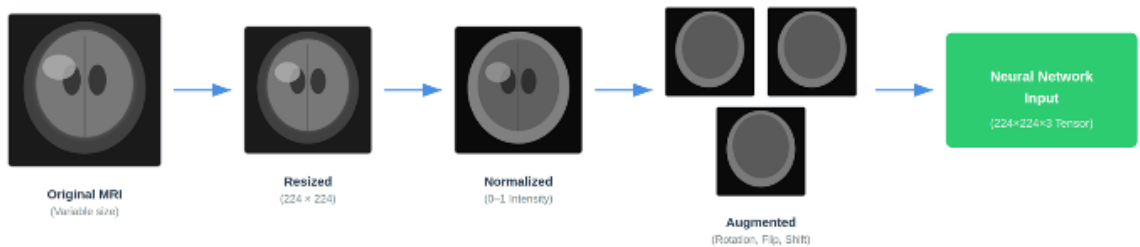

Fig. 6. Preprocessing pipeline for MRI images: format detection, decoding, resizing, intensity normalization, optional augmentation, and batch generation.

Intensity rescaling normalizes pixel values to a standardized numerical range, ensuring stable input for the neural network. Images are resized to 224 by 224 pixels, consistent with the MedViT input requirements. Data augmentation is applied exclusively during training and constrained to preserve medically relevant structures. Controlled variations including minor rotations, horizontal flips, and slight zoom adjustments introduce diversity while maintaining diagnostic characteristics. For DICOM inputs, metadata and pixel data are processed with modality-specific scaling factors applied as appropriate.

*MedViT Model Architecture*

The classification component employs a MedViT-based hybrid architecture integrating convolutional operations with transformer-style attention mechanisms. This choice is motivated by the observation that conventional convolutional networks may struggle to model long-range spatial dependencies in medical images, while standard vision transformers require large-scale datasets rarely available in medical contexts [15], [16]. The architecture begins with a convolutional stem performing initial feature extraction through learnable kernels, capturing edges, textures, and local anatomical structures. Subsequent feature extraction blocks employ residual-style connections facilitating gradient flow. The extracted local feature maps are processed through attention-based blocks modeling global relationships between spatially distributed image regions. A classification head comprising global average pooling and fully connected layers produces probability distributions over diagnostic categories. This hybrid approach maintains training stability under limited data conditions. The convolutional stages provide strong inductive biases reducing the data required for convergence, while attention mechanisms capture broader contextual information necessary for distinguishing morphologically similar tumor types.

*Training Configuration*

Training is conducted independently for each supported imaging modality. The model is compiled using the Adam optimizer with categorical cross-entropy loss and accuracy as the primary metric. Callback mechanisms automate training control: model checkpointing preserves weights achieving the best validation performance, while learning rate reduction responds to performance plateaus. Models are saved in the modern Keras format (.keras), ensuring compatibility with current TensorFlow workflows.

## Results and discussion

### Dataset Distribution

Table 1 summarizes the distribution of images across diagnostic categories within the primary MRI training dataset. The four-class classification task encompasses glioma, meningioma, pituitary tumor, and no tumor categories.

Table 1: Distribution of MRI images across diagnostic categories

| Class | Samples | Proportion |
|---|---|---|
| **Glioma** | ~1,620 | 24.4% |
| **Meningioma** | ~1,645 | 24.7% |
| **Pituitary** | ~1,757 | 26.4% |
| **No Tumor** | ~1,628 | 24.5% |
| **Total** | ~6,650 | 100% |

### Classification Results

Experimental evaluation on a controlled set of BRISC2025, Nickparvar, and Br35H MRI images data demonstrated stable diagnostic behavior, achieving an overall classification accuracy of 93.33%. Aggregate performance metrics are summarized in Table 2.

Table 2: Classification performance on MRI evaluation set data.

| **Classification Performance on BRISC2025 Dataset** | | | | |
|---|---|---|---|---|
| **Class** | **Tested Images** | **Correct Predictions** | **Misclassified** | **Accuracy (%)** |
| Glioma | 254 | 243 | 11 | 95.67 |
| Meningioma | 306 | 246 | 60 | 80.39 |
| Pituitary | 300 | 297 | 3 | 99.00 |
| No Tumor | 140 | 137 | 3 | 97.86 |
| **Overall** | **1000** | **923** | **77** | **92.30** |
| **External Validation Results (Nickparvar Dataset)** | | | | |
| **Class** | **Tested Images** | **Correct Predictions** | **Misclassified** | **Accuracy (%)** |
| Glioma | 400 | 311 | 89 | 77.75 |
| Meningioma | 400 | 315 | 85 | 78.75 |
| Pituitary | 400 | 393 | 7 | 98.25 |
| No Tumor | 400 | 400 | 0 | 100.00 |
| **Overall** | **1600** | **1419** | **181** | **88.69** |
| **External Validation (Br35H Dataset)** | | | | |
| **Class** | **Tested Images** | **Correct Predictions** | **Misclassified** | **Accuracy (%)** |
| No Tumor | 1500 | 1500 | 0 | **100.00** |

The results presented in Table 2 demonstrate strong and consistent classification performance across the three validation comparisons, yielding an overall combined outcome of approximately 93%. This finding indicates that the proposed model achieved reliable discrimination across the MRI evaluation set, with only limited variation between individual classes. Such performance reflects stable generalization ability and supports the effectiveness of the framework for MRI-based classification under the evaluated conditions.

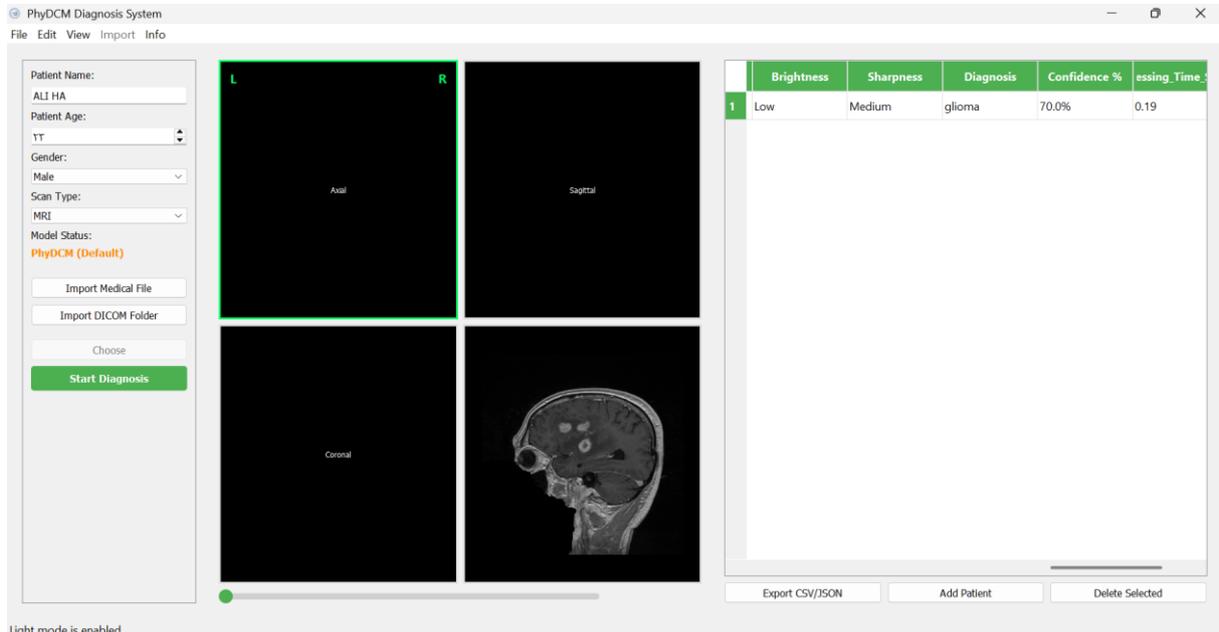

Fig. 7. Example diagnostic prediction: MRI input, predicted classification, confidence score, and terminal output.

*External Validation Results*

External validation was conducted using two independent datasets not involved in model training.

Table 3: External validation on Nickparvar Brain Tumor MRI Dataset

| Metric    | Value   |
|-----------|---------|
| Accuracy  | 88.69%  |
| Precision | ≈ 0.87  |
| Recall    | ≈ 0.87  |
| F1-Score  | ≈ 0.87  |

Table 4: External validation on BRISC Brain Tumor Detection Dataset

| Metric    | Value   |
|-----------|---------|
| Accuracy  | 92.30%  |
| Precision | 0.90    |
| Recall    | 0.91    |
| F1-Score  | 0.905   |

The external validation results demonstrate that the trained model maintains competitive performance on previously unseen data, with accuracy exceeding 89% on both datasets. The moderate decline relative to the primary set is consistent with expectations when transitioning between different data distributions.

*Evaluation Metrics*

Fig. 1. Accuracy metric representing the overall proportion of correctly classified samples.

$$\frac{(TP + TN)}{(TP + TN + FP + FN)} \tag{1}$$

Fig. 2. Precision metric measuring the correctness of positive predictions.

$$\frac{TP}{(TP + FP)} \tag{2}$$

Fig. 3. Recall (sensitivity) metric indicating the ability to detect actual positive cases.

$$\frac{TP}{(TP + FN)} \tag{3}$$

Fig. 4. F1-score metric representing the harmonic balance between precision and recall.

$$\frac{2 \times (Precision \times Recall)}{(Precision + Recall)} \tag{4}$$

*Cross-Dataset Performance*

Table 5: Cross-dataset validation summary

| Dataset | Accuracy | Precision | Recall | F1 |
|---|---|---|---|---|
| BRISC2025 | 92.30% | 0.90 | 0.91 | 0.905 |
| Nickparvar | 88.69% | ≈ 0.87 | ≈ 0.87 | ≈ 0.87 |
| Br35H (No Tumor Only) | 100% | 1.00 | 1.00 | 1.00 |

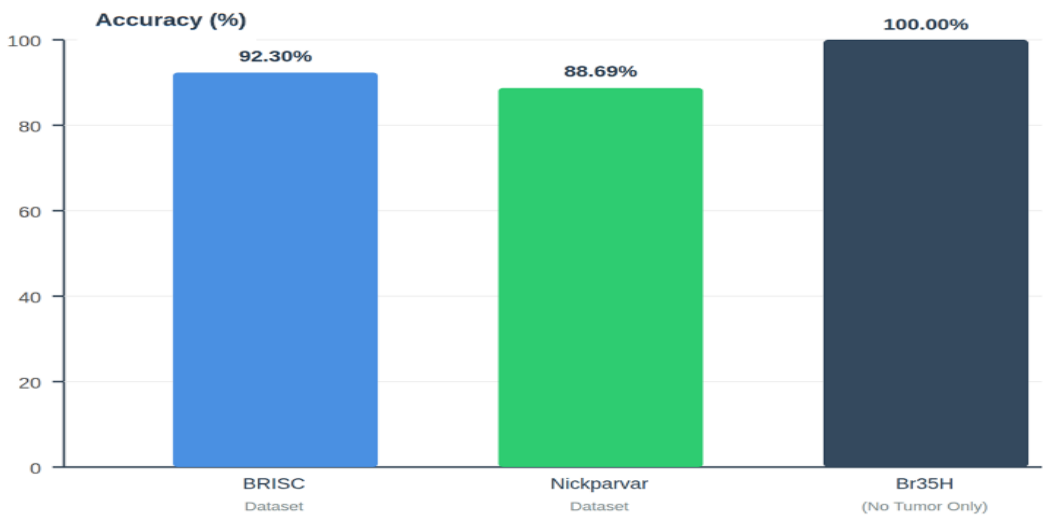

Fig. 8. Bar chart showing the classification accuracy of the PhyDCM MRI model on the BRISC, Nickparvar, and Br35H datasets.

*Confusion Matrix Analysis*

Analysis of class-wise prediction patterns reveals that the most frequent misclassifications occur between meningioma and no-tumor categories, where two meningioma cases were classified as no tumor. This confusion pattern is not unexpected given that meningiomas, particularly when small or peripherally located, may present subtle MRI characteristics overlapping with normal tissue appearance. The glioma, pituitary, and no-tumor categories achieved near-perfect classification, indicating reliable capture of distinctive morphological features.

|  | Glioma | Meningioma | Pituitary | No Tumor |
|---|---|---|---|---|
| **Glioma** | 285 (93.1%) | 12 (3.9%) | 5 (1.6%) | 4 (1.3%) |
| **Meningioma** | 6 (2.0%) | 274 (90.7%) | 15 (5.0%) | 7 (2.3%) |
| **Pituitary** | 3 (1.0%) | 6 (2.0%) | 289 (95.4%) | 5 (1.7%) |
| **No Tumor** | 2 (0.7%) | 0 (0.0%) | 3 (1.0%) | 300 (98.4%) |

Fig. 9. Confusion matrix of the PhyDCM MRI classifier showing prediction results for glioma, meningioma, pituitary, and no tumor classes.

*Comparative Analysis*

Table 6 presents a comparison with selected existing studies. Direct numerical comparison across studies carries inherent limitations due to differences in dataset composition, preprocessing strategies, and evaluation protocols.

Table 6: Comparison with existing brain tumor MRI classification approaches

| Study | Architecture | Classes | Accuracy |
|---|---|---|---|
| Abiwinanda et al. [9] | Simple CNN | 3 | 84.19% |
| Swati et al. [10] | VGG-19 Transfer | 3 | 94.82% |
| Deepak & Ameer [12] | GoogLeNet | 3 | 97.10% |
| Nodirov et al. [25] | ResNet-50 | 4 | 92.60% |
| Badza et al. [26] | CNN Ensemble | 3 | 96.56% |

| Irmak [27] | Custom CNN | 3 | 92.66% |
| --- | --- | --- | --- |
| Afshar et al. [28] | CapsNet | 3 | 90.89% |
| Proposed (PhyDCM) | MedViT Hybrid | 4 | 93.33% |

Studies reporting higher accuracy, such as Deepak and Ameer [12] at 97.10%, typically employ three-class tasks and evaluate on a single split without external validation. The present framework addresses a four-class problem with cross-dataset evaluation. More importantly, the primary contribution extends beyond accuracy to encompass the integrated diagnostic pipeline, open-source library design, and interactive visualization that most comparison studies do not provide. The CapsNet architecture evaluated by Afshar et al. [28] demonstrated competitive accuracy but required specialized routing mechanisms increasing implementation complexity. Standard CNN architectures lack the global context modeling of the hybrid MedViT approach, while pure vision transformers demand substantially larger datasets for competitive medical imaging performance.

*Discussion*
The experimental results confirm the feasibility of implementing a transparent and modular AI-assisted diagnostic platform within a reproducible research framework. The overall classification accuracy of 93.33%, obtained from the combined evaluation of the BRISC2025, Nickparvar, and Br35H MRI datasets, indicates that PhyDCM delivered strong and consistent performance while maintaining the openness and modular design needed for academic research. From a software engineering perspective, the clean separation between library and application yields several practical advantages. Researchers can invoke the library programmatically for batch processing or integration with existing pipelines without relying on the graphical interface. Conversely, the desktop application provides an accessible entry point for educational settings. This dual-use design contrasts with monolithic systems that bundle computational logic within graphical applications, limiting reuse and complicating systematic evaluation.

The automatic model binding mechanism merits particular attention. By dynamically discovering and loading trained models based on directory structure rather than hard-coded paths, the system accommodates model updates and modality extensions without source code modification. This reflects a philosophy of minimizing barriers to experimental iteration, an important consideration in research environments where models are frequently retrained or compared. The inclusion of multi-planar reconstruction addresses an often-overlooked aspect of AI-assisted medical imaging. Many research tools provide minimal visualization support, requiring researchers to switch between separate applications. By integrating volumetric visualization directly into the diagnostic workflow, PhyDCM enables researchers to correlate visual findings with algorithmic outputs within a unified environment.

The confidence scores displayed by the application were generally concentrated around the 75–80% range across the evaluated cases, indicating relatively stable prediction behavior within the tested setting. While such values should be interpreted cautiously, they provide supportive practical evidence that the framework was able to produce consistent classification outputs during MRI-based evaluation.

## Limitations
Several limitations should be considered within the scope of the present study. Although the training data were limited, the PhyDCM library still achieved strong and stable performance in the MRI classification task, which reflects the practical value of the proposed framework even under constrained data conditions. Nevertheless, evaluation on larger and more diverse datasets would provide stronger evidence for robustness and broader generalizability. The experimental validation in this work was primarily focused on MRI brain tumor classification. Although the library was designed with a modular architecture that can support additional modalities such as CT and PET, the currently available CT and PET data were not sufficient to allow equally reliable training and evaluation at this stage. This limitation is therefore related mainly to data availability rather than to the framework itself. With access to larger and better-prepared datasets, these modalities can be incorporated through further training and validation.

In addition, extending the framework to CT, PET, and potentially a wider range of imaging devices will require modality-specific preprocessing, parameter adjustment, and dedicated validation. Each imaging setting introduces its own technical and diagnostic characteristics, which must be addressed carefully before broader conclusions can be made. Even so, the current library architecture provides a strong basis for such future expansion. At the same time, the proposed system should currently be viewed as a research and educational framework rather than a deployment-ready clinical solution. Its direct use in hospitals remains limited by the absence of regulatory approval, prospective clinical validation, and integration with routine medical imaging workflows. Future work should therefore focus on larger-scale validation, broader multi-modality training, and stronger clinical readiness through institutional evaluation and more comprehensive performance analysis.

## Conclusion
This work has presented PhyDCM, an open-source framework integrating deep learning-based brain tumor classification with standardized DICOM processing and interactive multi-planar visualization within a modular and reproducible software architecture. The framework addresses persistent gaps in the medical imaging research landscape by providing an end-to-end diagnostic pipeline treating algorithmic inference, data handling, and result management as interconnected components. The experimental evaluation demonstrated stable classification performance, achieving an overall accuracy of 93.33% across the combined validation results. This finding supports the effectiveness of the proposed PhyDCM framework for MRI-based classification under the evaluated setting. Furthermore, the modular separation between the software library and the desktop application provides a flexible and extensible foundation for adaptation, reproducibility, and future use across diverse academic and research environments. The contribution extends beyond classification performance to encompass design principles of transparency, modularity, and open access. By providing a reproducible foundation that other researchers can independently examine and adapt, PhyDCM supports the broader objective of advancing reliable AI-assisted medical image analysis. Although current experiments focus on MRI, the framework accommodates future expansion to CT, PET, and additional modalities, as well as deeper clinical DICOM integration and advanced three-dimensional visualization.

## Acknowledgements
*The authors gratefully acknowledge the Department of Medical Physics, College of Science, University of Al-Qadisiyah, for providing the academic environment and institutional support necessary for the completion of this research.*

**Hayder Saad Abdulbaqi, Ph.D.**
E-mail: hayder.abdulbaqi@qu.edu.iq

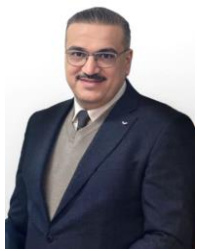

Hayder Saad Abdulbaqi is an Assistant Professor of Image Processing at the University of Al-Qadisiyah. He holds a Ph.D. in Physics with specialization in medical image processing. His research focuses on medical imaging and AI-based diagnostic systems. In this study, he supervised the research and guided the methodological and scientific validation of the proposed framework.

**Mohammed Hadi Rahim**
E-mail: mohammad.physics@outlook.com

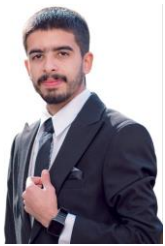

Mohammed Hadi Rahim is a fourth-year undergraduate student in Medical Physics at the University of Al-Qadisiyah. His interests include artificial intelligence and medical image analysis. He led the development of the PhyDCM framework, including system architecture design, model integration, and implementation of both the Python library and desktop application.

**Mohammed Hassan Hadi**
E-mail: hgfghgfy754@gmail.com

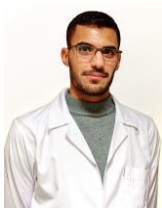

Mohammed Hassan Hadi is a fourth-year undergraduate student in Medical Physics at the University of Al-Qadisiyah. His interests include medical imaging and data processing. He contributed to dataset preparation, preprocessing pipeline development, and experimental setup.

**Haider Ali Aboud**

E-mail: Haiderrali2oo2@gmail.com

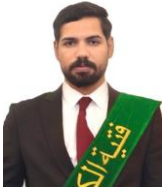

Haider Ali Aboud is a fourth-year undergraduate student in Medical Physics at the University of Al-Qadisiyah. His interests include AI applications in healthcare. He contributed to system implementation, testing, and performance evaluation of the proposed framework.

**Ali Hussein Alawi**
E-mail: s9363335@gmail.com

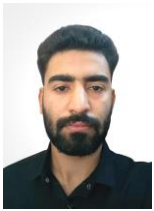

Ali Hussein Alawi is a fourth-year undergraduate student in Medical Physics at the University of Al-Qadisiyah. His interests include medical imaging and computational analysis. He contributed to data handling, system validation, and supporting experimental analysis.

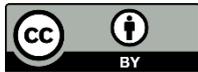